\documentclass[journal,twocolumn]{IEEEtran}

\usepackage{threeparttable}
\usepackage{pifont}
\usepackage{flushend}

\usepackage{currfile-abspath} 
\getmainfile
\getabspath{\themainfile}

\usepackage{xcolor} 

\usepackage{siunitx}
\ifCLASSINFOpdf
  \usepackage[pdftex]{graphicx}
  \graphicspath{../imgs/}
  \DeclareGraphicsExtensions{.pdf,.png,.jpg}
\else
\fi

\usepackage{amsmath}
\usepackage{booktabs}
\usepackage{multirow}

\ifCLASSOPTIONcompsoc
  \usepackage[caption=false,font=normalsize,labelfont=sf,textfont=sf]{subfig}
\else
  \usepackage[caption=false,font=footnotesize]{subfig}
\fi

\usepackage{url}

\begin{document}

\title{Benchmarking Scientific Image Forgery Detectors}

\author{ \IEEEauthorblockN{João~P.~Cardenuto}, 
        Anderson~Rocha,~\IEEEmembership{Senior~Member,~IEEE}
        
        \IEEEauthorblockA{RECOD~Lab.,~Institute~of~Computing,~University~of~Campinas}
        \\Email:~\{phillipe.cardenuto,~anderson.rocha\}@ic.unicamp.br
        }

\maketitle

\begin{abstract}
 The scientific image integrity area presents a challenging research bottleneck, the lack of available datasets to design and evaluate forensic techniques. Its data sensitivity creates a legal hurdle that prevents one to rely on real tampered cases to build any sort of accessible forensic benchmark. To mitigate this bottleneck, we present an extendable open-source library that reproduces the most common image forgery operations reported by the research integrity community: duplication, retouching, and cleaning. Using this library and realistic scientific images, we create a large scientific forgery image benchmark (39,423 images) with an enriched ground-truth. In addition, concerned about the high number of retracted papers due to image duplication, this work evaluates the state-of-the-art copy-move detection methods in the proposed dataset, using a new metric that asserts consistent match detection between the source and the copied region. The dataset and source-code will be freely available upon acceptance of the paper.
\end{abstract}


\section{Introduction}
\label{sec:intro}
Integrity researchers have been reporting a threat to scientific image integrity for a long time\cite{Krueger2002,rossner2004what,Gilbert2009, Bik2016prevalence}. This improbity has achieved even areas like cancer \cite{Schiermeier2002}, or more dramatically used on 'paper mills' services \cite{Chawla2020}.

Although this serious problem has been an urge in the scientific community, to the best of our knowledge, few forensics works are dedicated to this topic. So far, no rich annotated forensic benchmark containing scientific tampering images was published. We believe that a large dataset would foster the forensic community to work more actively in this subject and assist state-of-the-art forensic techniques that might require large training datasets.

\begin{figure}[!t]
\centering
\includegraphics[width=3.5in]{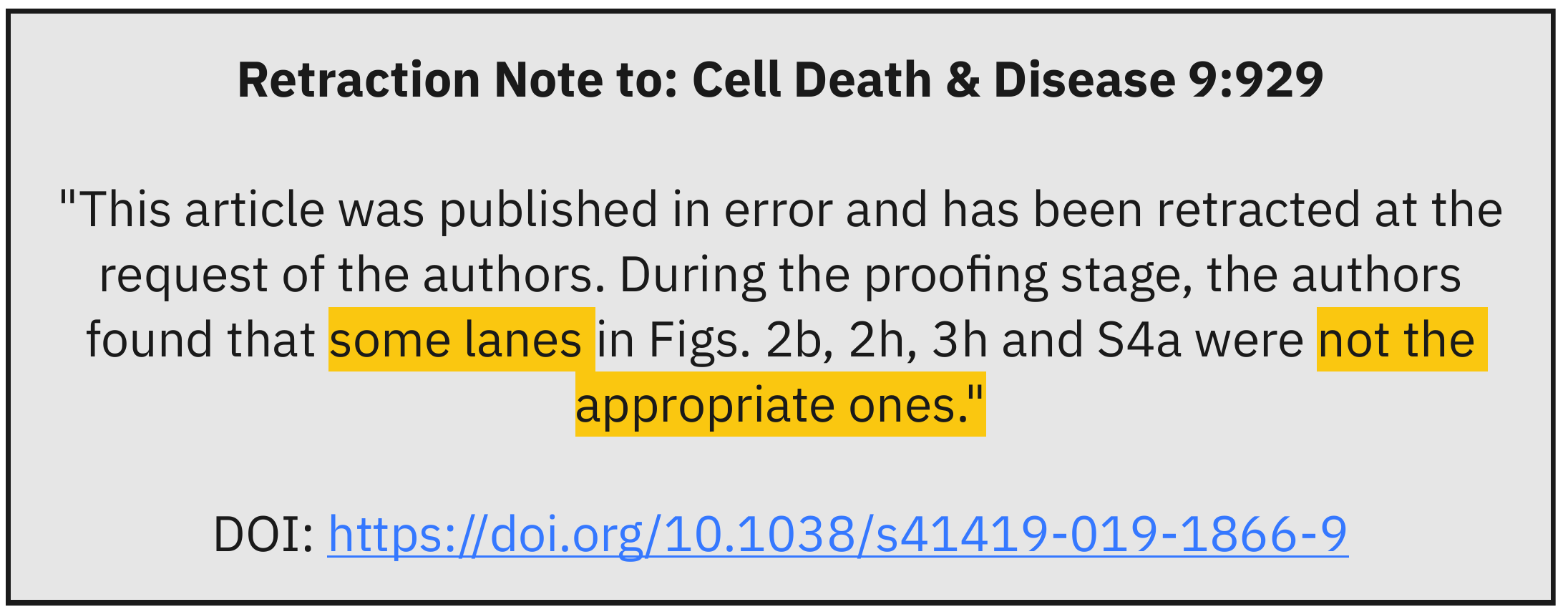}
\caption{Example of a retraction note extracted from Cell Death \& Diseases (https://www.nature.com/articles/s41419-019-1866-9, Last access May, 2021). The highlighted words in yellow `some lanes' and `not the appropriate ones' illustrate inaccurate regions and ambiguous causes of the retraction.}
\label{fig:retractio-notice}
\end{figure}

In this sense, we tried to collect known doctored scientific images, but we faced two main issues that make us avoid these methodology: legal and practical.
To publish a dataset with real tampering cases, we would have to face copyrights and legal aspects of pointing third-party works that were retracted due to suspicious manipulation. Even if we decided to manage this legal aspect, we had to be guided by a retraction notice relative to the issued images. However, after reading several retraction notices, we realized that many of them are not precise enough to pinpoint the issued region's at a pixel level, which would not lead to an accurate ground-truth. The example of a real retraction notice due to an honest error, depicted in Fig. \ref{fig:retractio-notice}, shows this inaccuracy, in which the highlighted words `some lanes' and `not the appropriate ones' translate to an ambiguous region and cause.


On the other hand, we notice that a representative number of retracted images were due to duplication and basic image processing operations that could be automatically created.
Therefore, this work presents the RECOD Scientific Image Integrity Library (\textsc{RSIIL}) that enables creating a synthetic scientific image tampering dataset with enriched pixel-wise ground-truth and without any associated legal issue. 
With this library, we created the RECOD Scientific Image Integrity Dataset (\textsc{RSIID}) with the most common image operations reported by scientific integrity researchers \cite{rossner2004what,Bik2016prevalence}. To create this benchmark, we doctored 2,923 figures from creative common sources resulting in 39,423 tampered figures (26,496 for training and 12,927 for testing).
In addition, we propose a new metric to evaluate copy-move forgery detection dedicated to scientific images using an enriched ground-truth map, that assert a consistent detection match between the cloned region and its source. Finally, using this new dataset and metric, we evaluate the performance of the state-of-the-art copy-move forgery detection \cite{cozzolino2015efficient,Christlein2012,wu2018eccv}, establishing a baseline and setting the ground for any future investigation.

We organize the remaining of this paper into seven sections: Section~\ref{sec:relwork} presents related work while Section~\ref{sec:forgery-lib} details the proposed library, \textsc{RSIIL}. Section~\ref{sec:forgery-data} presents the dataset \textsc{RSIID} while Section~\ref{sec:metric} brings out a new evaluation metric aimed at a more consistent copy-move detection evaluation. Section~\ref{sec:eval} presents an analysis of state-of-the-art copy-move forgery detectors on the proposed dataset setting the ground for future research while Section~\ref{sec:conclusions} presents the conclusions and future work directions.

\section{Related Work}
\label{sec:relwork}

To the best of our knowledge, few works try to design a tampering benchmark focused on scientific images. So far, we were only able to find two works that address scientific integrity image datasets.
The first one is from Xiang and Acuna \cite{xiang2020scientific}, which created a synthetic tampering dataset of scientific images from the web. They doctored microscopy and western blot images using three types of manipulations that they claim to be the common cause of problems in scientific papers: cleaning of an image region with a single color or noise (Cleaning); copying an alien content region into the image (Splicing); and applying visual adjustments in the image content (Retouching). Their dataset contains 747 manually manipulated scientific images, of which 616 are dedicated to Removal. As we were only able to find the pre-print version of~\cite{xiang2020scientific}, we could not find any released data. Due to this, the quality of their manipulations and the dataset license is still unclear. Despite the authors manually constructed the dataset to create a more realistic scenario, their dataset is still limited to a small size that might not represent the diversity of scientific images. Besides this, the dataset is highly concentrated on Cleaning, preventing one to properly evaluate the robustness of a forensic method among all modalities.

The second one is the work of Koker et al. \cite{koker2021icpr}, named as Bio-Image Near-Duplicate Examples Repository (BINDER), which have the pioneering idea of using legal issue-less scientific images for an integrity dataset. This dataset is limited to finding near-duplicate images, aiming to find image re-use across scientific publications. Their dataset has 10,179 non-overlapping patches tiled in
$256\times256$ or $128\times128$ pixels. To create their dataset, they gathered microscopy images from the following public repositories: NYU Mouse Embryo Tracking Database\footnote{http://celltracking.bio.nyu.edu (Last access May, 2021} (METD), the Broad Bioimage Benchmark Collection\footnote{https://bbbc.broadinstitute.org (Last access May, 2021)} (BBBC), the Adiposoft Image Dataset\footnote{https://imagej.net/Adiposoft (Last access May, 2021)} (AID), and the Open Microscopy Image Data Resource\footnote{https://idr.openmicroscopy.org (Last access May, 2021)} (IDR). Besides, they also applied some geometric, brightness/contrast, and compression transformations on some images. However, their dataset is still not as realistic as the figures presented in scientific publications.
Despite scientific images often embed graphical elements and captions, hampering to detect re-use, the authors did not add these elements to the images. In addition, they did not apply any local tampering (region-level), which is also a typical manipulation in inappropriate image re-use \cite{Bik2016prevalence}.

In addition to these works, we also found two scientific integrity initiatives that collect real cases of retracted papers.
The first is the Retraction Watch Database\footnote{http://retractiondatabase.org (Last access May, 2021)} maintained by the Retraction Watch\footnote{A non-profit organization affiliated with the Center for Scientific Integrity and dedicated to report and discuss cases of retracted papers and related issues.}. This database has more than 20,000 metadata of retracted, corrected, or concerned papers. The metadata presents the paper's title, retraction reason, authors, and Digital Object Identifier (DOI), among other fields. Although this database is not dedicated to image integrity issues, it is possible to filter the retracted papers to this category. However, only the paper's metadata will be retrieved – due to legal aspects; it is not possible to retrieve the articles PDF, Figures, or Retraction notice –, which is a drawback of this database.

The second is the HEADT Centre Image Integrity Database\footnote{https://headt.eu/Image-Integrity-Database (Last access May, 2021)}, an initiative focused on researchers and developers working on scientific image manipulation detection. Their database contains more than 500 images’ metadata from retracted papers due to image manipulation. In addition to the basic information of a paper (Title, Authors, Publisher, Journal), they also added a text description of each manipulation, including the figure’s panel in which the manipulation occurs and its category (e.g., copy-move). This text description is based on the retraction notice associated with the figure; therefore, some text also presents ambiguity as depicted in the Figure \ref{fig:retractio-notice}. Despite this text description, we could not find any manipulation map at pixel level in this dataset, which we believe is needed to evaluate a detection method properly.

\section{RECOD Scientific Image Integrity Library - RSIIL}
\label{sec:forgery-lib}

    

Before working with synthetic data, we tried to gather real-world problematic scientific images. To avoid any bias from our side, we relied upon retracted papers due to image problems given that they have a retraction notice resultant of an integrity investigation. However, to publish an accessible benchmark for forensic research, we possibly would have to deal with some legal aspects (e.g., figure copyright and causing possible defamation to someone).

As reported by Adam Marcus \cite{marcus_2019}, a retracted paper could make their authors feel their reputation harmed and make them sue journals for defamation. Azoulay et al. \cite{ azoulay2017career} also indicate that retraction due to misconduct --–which are the most important papers to be included in a forensic benchmark--- has a significant reputation penalty to their authors. Even co-authors that might not be involved in the image manipulation, who already suffered severe consequences \cite{mongeon2013collective}, could be affected by such benchmark since it would promote their association with the retracted paper.

Besides this legal aspect, we also faced some practical issues regarding data annotation. When manually annotating the problematic figures’ regions following their retraction notice, we experienced an absence of standard, including vague sentences (as illustrated in Figure \ref{fig:retractio-notice}), resulting on unreliable ground-truths.

Because of these issues, we decide to avoid using real-world scientific problematic image and create a photorealistic dataset using the library introduced in this section. Thus, this Section presents the types of forgeries implemented in the library (Sub-Section A), explains how the library mimics realistic figures as they usually are presented in scientific documents (Sub-Section B), and addresses the manipulation ground-truth (Sub-Section C). Finally, the section also discusses how the proposed library is amenable to extensions of new image manipulations types (Sub-Section D).

\subsection{Library functionalities}

The goal of the library is to implement the most common image manipulations reported in the scientific community.
Although we are aware of the possibilities of more complexity tools for image manipulation, for example, the creation of scientific images using artificial intelligence (AI) algorithms \cite{qi2020emerging}, we suspect that these tools are not vastly used yet due to their complexities. 
Therefore, while designing the library, we adopt the forgery function based on the most common image processing operation accessible for a non-expert in AI or Computer Vision. We also design each block of the library to allow it to be extended to other more complex operations in the future.
 
 Following the research from Bik et al. \cite{Bik2016prevalence} and Rossner and Yamada \cite{rossner2004what}, we selected three main types of manipulations that can be recreated using common image processing software (e.g., Adobe Photoshop):

\begin{enumerate}
    \item \textbf{Retouching}: The process of image beautification leading to an experiment misreading. This modality implements contrast, brightness, and blurring adjustments that highlight or obfuscate an image region. 
    Figure \ref{fig:retouching} depicts an image that we applied retouching with our library. In Figure \ref{fig:blur-manipulated}, we used a Gaussian filter within the selected objects to obfuscate its content. Figure \ref{fig:contrast-manipulated} illustrates an image with contrast and brightness adjustment, in which the method changes the selected object pixels intensity to cause an experimental misreading.

    \begin{figure}
    \centering
     \subfloat[Original\label{fig:blur-original}]{%
    \includegraphics[width=0.15\textwidth]{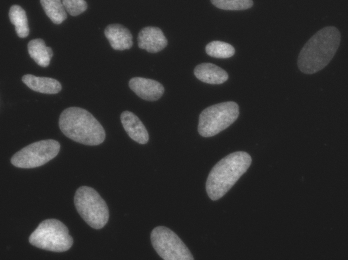} }
    \subfloat[Manipulated\label{fig:blur-manipulated}]{%
    \includegraphics[width=0.15\textwidth]{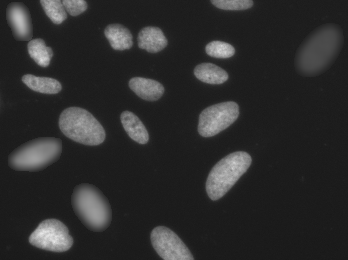} }
    \subfloat[Manipulation Map\label{fig:blur-gt}]{%
    \includegraphics[width=0.15\textwidth]{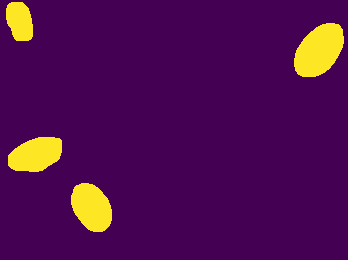} }
    \hfill
    \subfloat[Original\label{fig:contrast-original}]{%
    \includegraphics[width=0.15\textwidth]{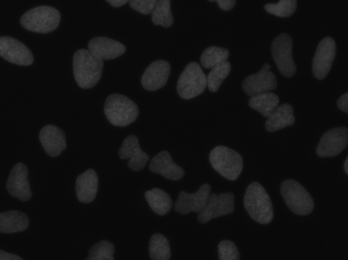} }
    \subfloat[Manipulated\label{fig:contrast-manipulated}]{%
    \includegraphics[width=0.15\textwidth]{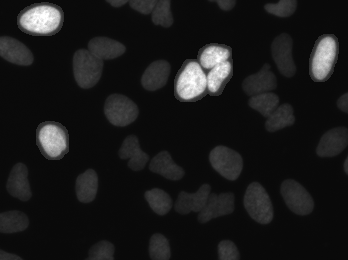} }
    \subfloat[Manipulation Map\label{fig:contrast-gt}]{%
    \includegraphics[width=0.15\textwidth]{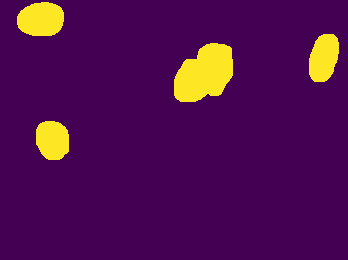} }
    
    \caption{Example of image Retouching forgery implemented in the library. (a) and (d) are original images without any manipulation; (b) is the manipulated version of (a) using blurring retouching; (e) is the brightness/contrast manipulated version of (d); (c) and (f) are the ground-truth map that indicate the manipulated regions of (b) and (e).}
    \label{fig:retouching}
     \end{figure}
    
  \item \textbf{Cleaning}: The result of obfuscating a foreground object using a background region. For this modality, we use inpainting and a brute-force routine. For the inpainting, we use the method of Criminisi et al. \cite{criminisi2004region} implemented by Moura.\footnote{Code available at https://github.com/igorcmoura/inpaint-object-remover. (Last access March, 2021)} For the brute-force routine, we develop an in-house method to mimic the forgery procedure of a person seeking to cover an object using the background. To implement this routine, we select a foreground object $\mathcal{FO}$ ; then, using brute force, we fit  $\mathcal{FO}$ on a background region  $\mathcal{BR}$ that has the most similar color histogram of this object; finally, we copy $\mathcal{BR}$ into $\mathcal{FO}$ and blur the border of $\mathcal{FO}$, smoothing (feather edges) the difference from the copied $\mathcal{BR}$ and the neighborhood of $\mathcal{FO}$.
  Figure \ref{fig:inpainting-manipulated} depicts the result of inpainting on the top-right cell of the image, and Figure \ref{fig:cleaning-manipulated} depicts the result of the brute-force routine.
  
  \begin{figure}
    \centering
     \subfloat[Original\label{fig:inpainting-original}]{%
    \includegraphics[width=0.13\textwidth]{imgs/cleaning/inpainting_original.png} }
    \subfloat[Manipulated\label{fig:inpainting-manipulated}]{%
    \includegraphics[width=0.13\textwidth]{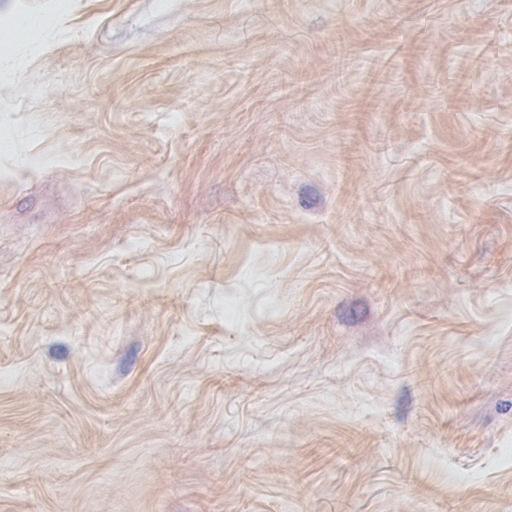} }
    \subfloat[Manipulation Map\label{fig:inpainting-gt}]{%
    \includegraphics[width=0.13\textwidth]{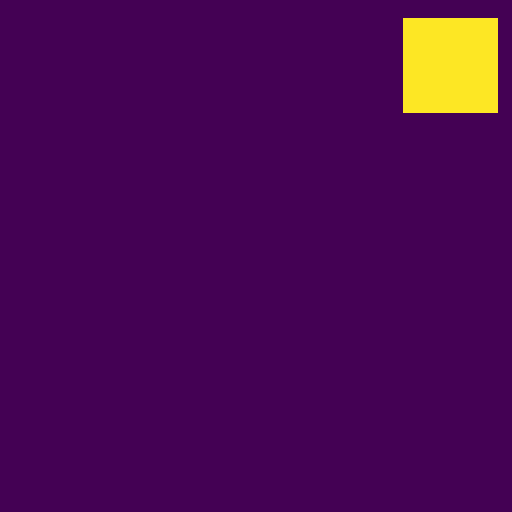} }
    \hfill
    \subfloat[Original\label{fig:cleaning-original}]{
    \includegraphics[width=0.13\textwidth]{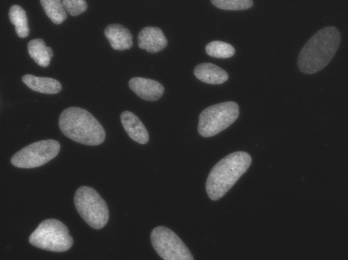}}
    \subfloat[Manipulated\label{fig:cleaning-manipulated}]{
    \includegraphics[width=0.13\textwidth]{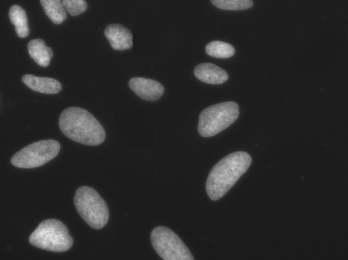}}
    \hfill
    \subfloat[Foreground Map\label{fig:cleaning-fg-gt}]{%
    \includegraphics[width=0.13\textwidth]{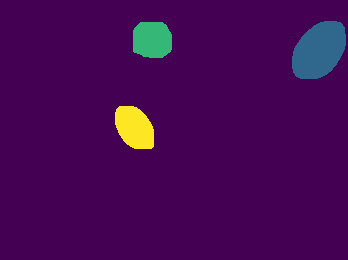} }
     \subfloat[Background Map\label{fig:cleaning-bg-gt}]{%
    \includegraphics[width=0.13\textwidth]{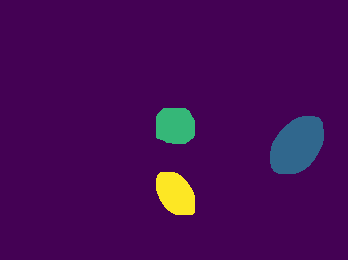} }
    
    \caption{Example of image Cleaning forgery implemented in the library. (a)-(c) depict the inpainting method of \cite{criminisi2004region} added to the library. (d)-(g) depict the Brute-Force cleaning routine. (g) indicates the background regions of (d) selected to cover (clean) the cells indicated by (f). Each  color in (f) and (g) represent a different ID that helps to track the regions involved in the forgery.}
    \label{fig:cleaning}
     \end{figure}

   \item \textbf{Duplication}: The action of copying and pasting a region of an image within the same or another image, using or not post-processing operations. Note that this definition includes both copy-move and splicing.
   We organized this category into three sub-categories: 
   
   \begin{enumerate}
       \item \textbf{Copy-Move Forgery}: Duplication of a region within the same image using geometric transformations (translation, rotation, flip, and scaling) and post-processing (e.g., retouching). All transformations can be combined with another. Due to the intrinsic result of scaling, we always combined it with another operation, otherwise it would cover the source object region.
       Besides these transformations, we also implemented a random object-to-background copy-move (that we named Random). This routine copies a random object $\mathcal{RO}$ to a background region $\mathcal{BR}$ that has the same shape as $\mathcal{RO}$.
       \item \textbf{ Overlap Forgery}: Creation of two images with an overlap region from a single one. From a source image $\mathcal{I}$, we select different regions of $\mathcal{I}$ that share an overlap area to create two images from these regions. Any of these new regions can suffer post-processing to obfuscate its source. Figure \ref{fig:overlap} depicts the creation of an image with an overlap area.
       \item \textbf{Splicing}: Creation of an image composition that uses a donor figure's elements into a host one. Figure \ref{fig:splicing} depicts an Splicing forgery.
       
   \end{enumerate}
 \end{enumerate}

\begin{figure}
        \centering
        \subfloat[Translation\label{fig:cmfd-translation}]{%
        \includegraphics[width=0.1\textwidth]{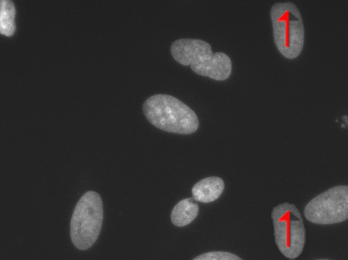}
        }
        \subfloat[Rotation\label{fig:cmfd-rotation}]{%
        \includegraphics[width=0.1\textwidth]{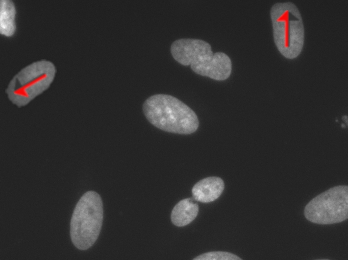} }
         \subfloat[Flip\label{fig:cmfd-flip}]{%
        \includegraphics[width=0.1\textwidth]{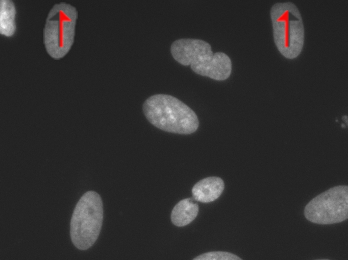} 
        }
         \subfloat[Scaling\label{fig:cmfd-scale}]{%
        \includegraphics[width=0.1\textwidth]{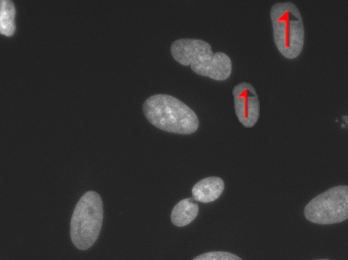} 
        }
        \hfill
         \subfloat[Original\label{fig:cmfd-original}]{%
        \includegraphics[width=0.15\textwidth]{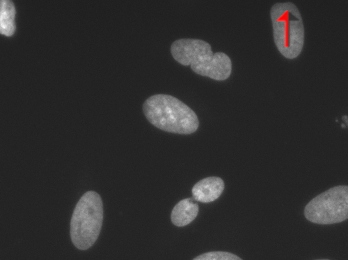} }
        \caption{Example of Copy-Move Forgery implemented in the library. The object of image (e) containing an arrow is duplicated with (a) translation, (b) rotation, (c) flip, and (d) scaling and pasted within the same image.}
        \label{fig:cmfd}
\end{figure}

\begin{figure}
        \subfloat[Source Image\label{fig:overlap-src}]{%
        \includegraphics[width=0.17\textwidth]{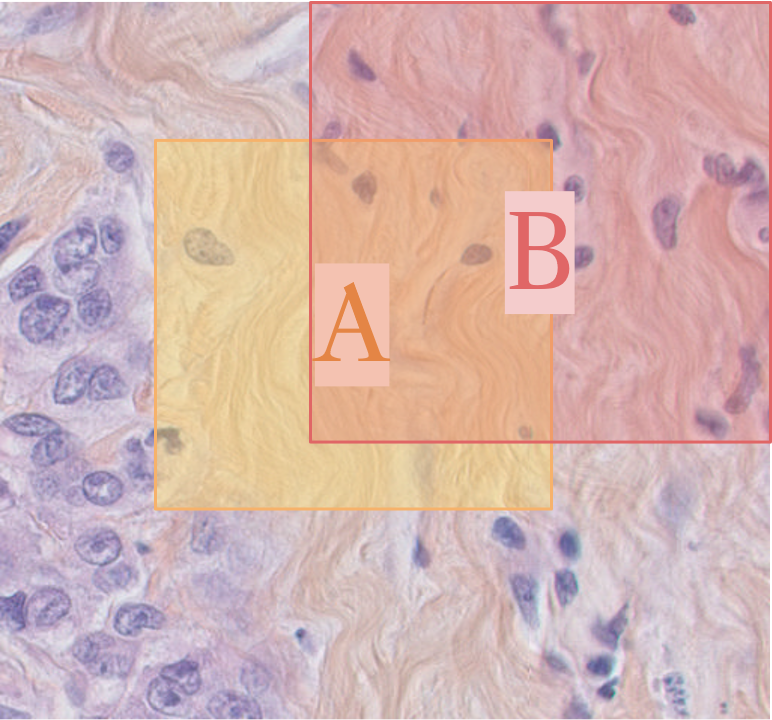}
        }
        \hfill
        \centering
         \subfloat[Region A\label{fig:overlap-A}]{%
        \includegraphics[width=0.15\textwidth]{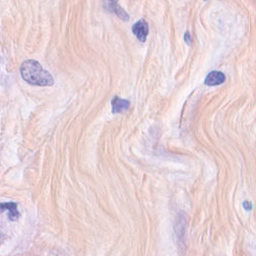} }
         \subfloat[Region B\label{fig:overlap-B}]{%
        \includegraphics[width=0.15\textwidth]{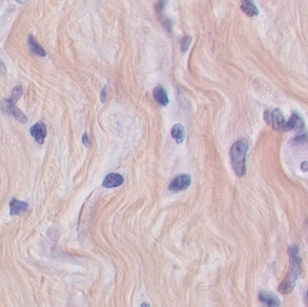} 
        }
        \caption{Example of Overlap forgery included in the library. (a) represent a  source image that is divided in overlapping regions A and B, and then presented as unique images in (b) and (c). The region A (b) suffer a post-processing brightness adjustment to make harder to compare with region B (c)}
        \label{fig:overlap}
\end{figure}

\begin{figure}
        \centering
         \subfloat[Donor\label{fig:donor}]{%
        \includegraphics[width=0.15\textwidth]{imgs/duplication/duplication_arrow.png} }
         \subfloat[Host\label{fig:host}]{%
        \includegraphics[width=0.15\textwidth]{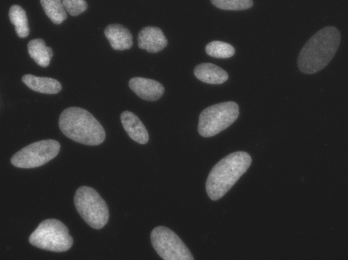} 
        }
        \hfill
        \subfloat[Splicing Result\label{fig:splicing-src}]{%
        \includegraphics[width=0.17\textwidth]{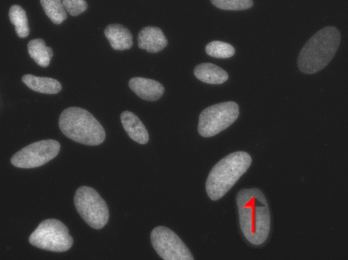}
        }
        \caption{Example of Splicing forgery function included in the library. The object highlighted with a red arrow from the donor image (a) is placed in a background region of the host image (b) resulting in (c).  }
        \label{fig:splicing}
\end{figure}


Despite all cases are generated without any human interaction, the result images may confuse even an attentive person.  To produce forgeries as realistic as possible, some functions from the library require as input an object map (segmentation map). The object map locates each object inside the image and assists a forgery function to execute the falsification more likely as a human would do.

\subsection{Realistic Scientific Figures}
\label{subsec:realist-scifigs}
As our key objective is to create scientific figures, we include two features (frequently present in such figures) in the library: captions/indicative letters and compound figures.

\begin{enumerate}
    \item   \textbf{Caption/indicative letters}: Scientific figures often present indicative letters or captions that overlay the image's content. As a result, this overlay is a splicing operation between a letter or a word within the experiment image that could raise a false alarm during forgery detection. Therefore, we add to the library the possibility to mimic this overlap behavior as it appears on scientific papers.
    We include three different levels of indicative verbosity. Level 1 includes only indicative letters around each panel of the figure. Level 2 includes the features of Level 1 and a random word around each panel. Level 3 includes all features from Level 2 and an indicative letter inside each panel. Figure \ref{fig:verbosity-figs} depicts all these levels of verbosity.
    
    \begin{figure}
        \centering
        \includegraphics[width=0.3\textwidth]{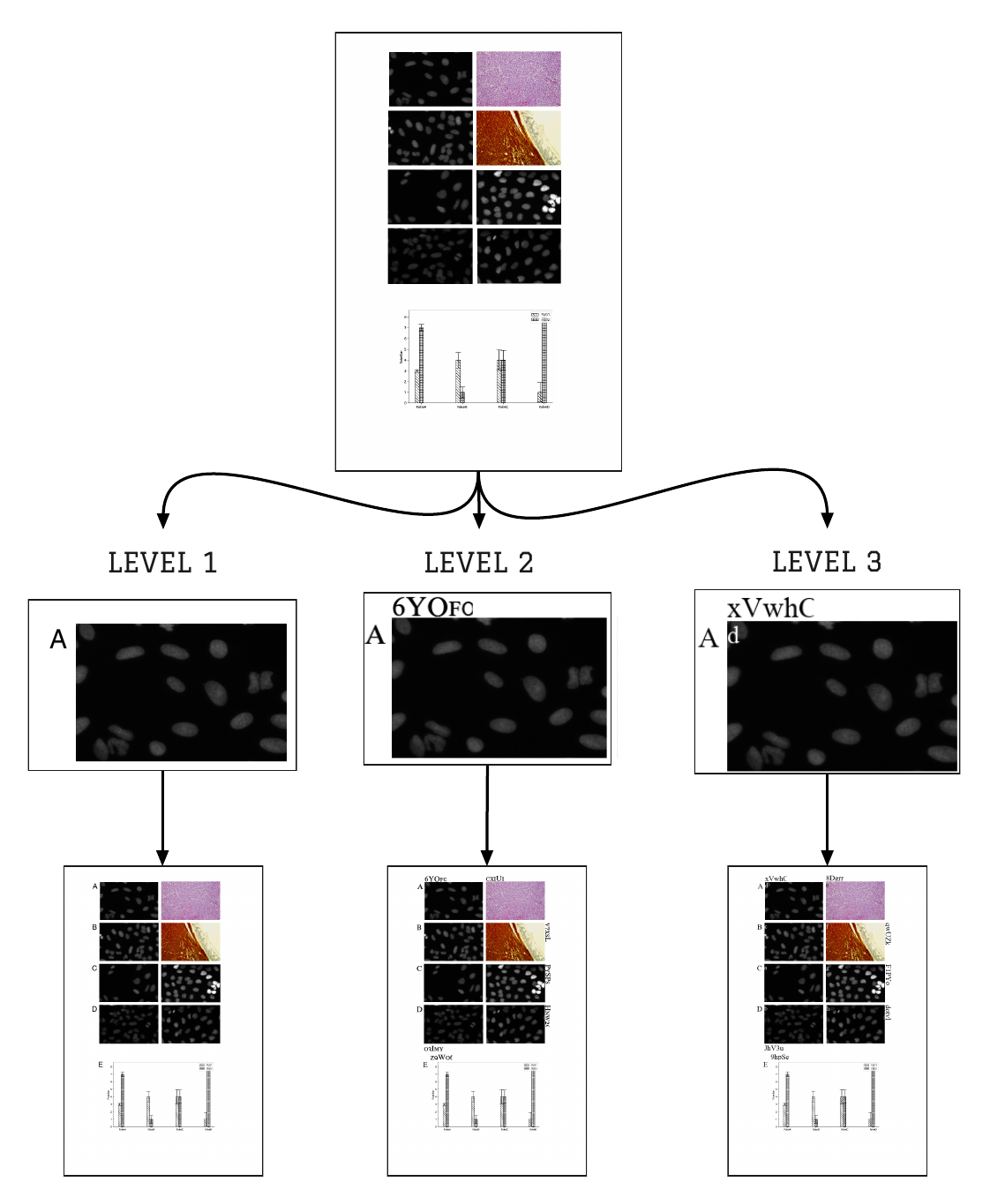}
        \caption{Example of a compound figure with different levels of indicative letters verbosity. From top to bottom, a compound figure without any indicative letter receives different levels of indicative verbosity, depicted by one of its panels; at the bottom, the result figures for each level.}
        \label{fig:verbosity-figs}
\end{figure}

    \item \label{subsec:compound-fig} \textbf{Compound figures}: A \textit{Compound} figure is a composition of multiple images that are organized in panels. These figures usually appear in articles to represent an overview of an experiment. To avoid creating unrealistic compound figures, we make use of figure templates based on real cases. These templates are image masks that can inform each panel's location to the method, as well as their type (e.g., graphs, photos).
\end{enumerate}
To create \textit{Compound} figures, we implemented a routine that has as input a set of realistic compound figures templates $\mathcal{T}$, a dataset of scientific images $\mathcal{D}$ (to be included in the compound figure), a source image $\mathcal{S}$ ($\mathcal{S}$ is not in $\mathcal{D}$), and a forgery function $f$ (to be applied in $\mathcal{S}$). Figure  \ref{fig:compound-creation-pipeline} illustrates this routine.
Thus, the method selects a template $t$ from $\mathcal{T}$ with at least one panel whose aspect ratio is similar to the aspect ratio of $\mathcal{S}$. Then,  the routine applies the forgery $f$ in $\mathcal{S}$, creating $\mathcal{S}_f$ (a forged version of $\mathcal{S}$). Later, a figure with the same size of $t$ is created, and $\mathcal{S}_f$ is resized and placed in the panel of $t$ with the most similar aspect ratio of $\mathcal{S}_f$. Finally, all other panels of $t$ are filled with different images from $\mathcal{D}$ that have similar aspect ratio to those panels or with fake graphics.

\begin{figure}
        \centering
        \includegraphics[width=0.4\textwidth]{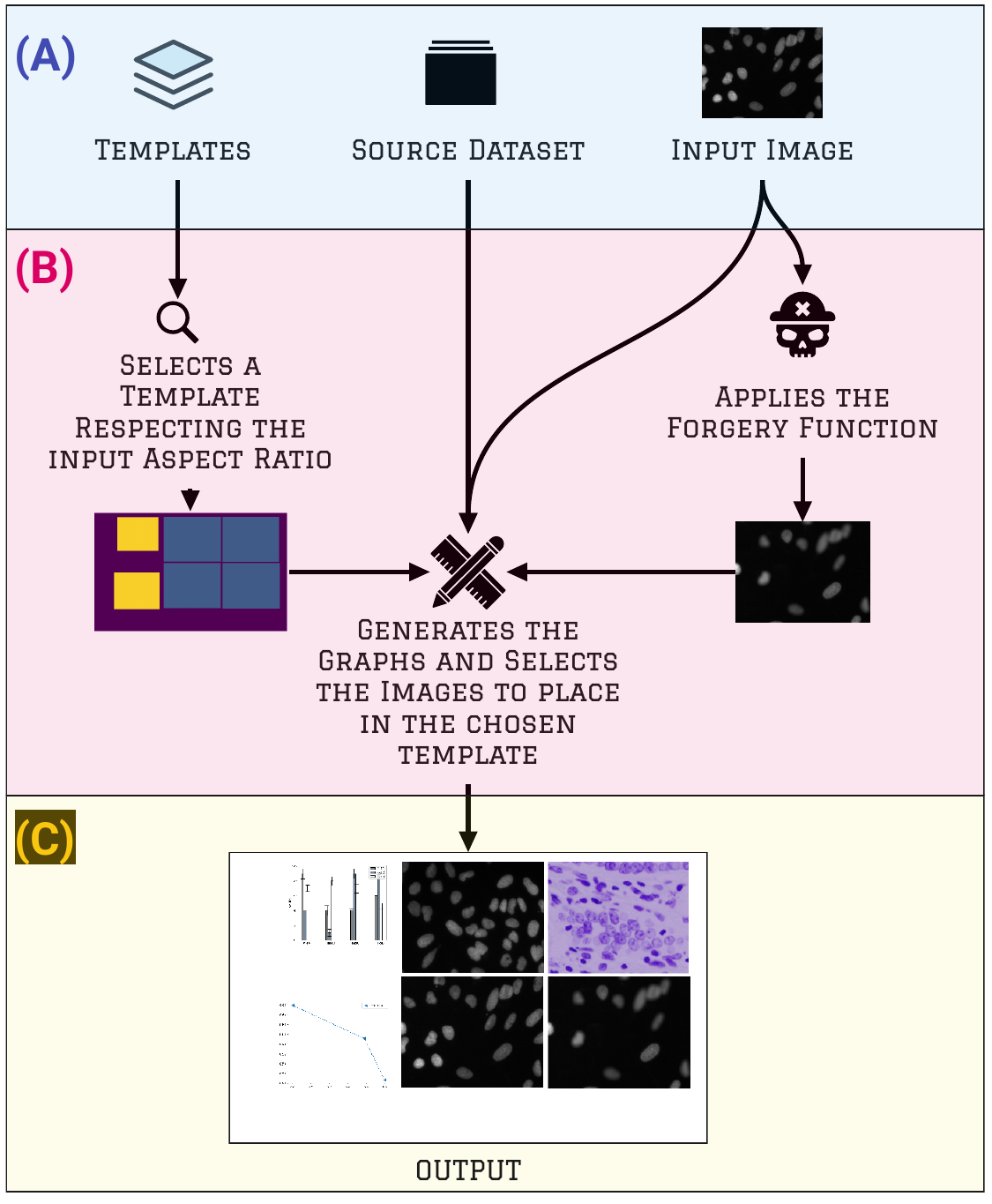}
        \caption{Pipeline of \textit{Compound} figure creation. (a) Method's Input: From left to right, set of \textit{Compound} Figures templates; scientific source image dataset; and input image with the chosen forgery function. (b) Method's Operations: Selects a template based on the aspect ratio of the input image; then, retrieves all images from the source dataset that fit the chosen template; later, creates Fake graphs (if indicated by the template); then, applies the forgery function in the input; and, finally, place all figure elements in the \textit{Compound} figure. (c) The output figure}
        \label{fig:compound-creation-pipeline}
\end{figure}



\subsection{Data Annotation}
In spite of the importance of reliable ground-truth to evaluate a forensic method, to the best of our knowledge, there is no scientific forgery image dataset that presents an enriched ground-truth. 
Hence, all tampered operations implemented in the library provide detailed maps to indicate the manipulated regions. Each object involved in the tampering operation is indicated with a different ID in the ground-truth, which helps pinpoint the object's exact location before and after the forgery, as depicted in Figure \ref{fig:cleaning-bg-gt}.
The library also enables the creation of a JSON file containing metadata related to the forgery. This metadata includes the source images, the method and arguments used, and the location of each panel inside the \textit{Compound} figure. As the metadata includes the source images and the forgery methods applied, one can evaluate provenance analysis \cite{moreira2018image} using these information as reference.

\subsection{Library Extension}

Given that scientific image tampering improves over time to convince even researchers \cite{qi2020emerging}, the benchmark of tampering detection also should include cutting-edge forgery techniques. In this sense, to facilitate the inclusion of new manipulation in RSIIL, we implemented a high-level routine that receives as one of its arguments a forgery function and applies it to an image. This routine is responsible for regulating the application of any new manipulation, asserting its guidelines to the ground-truth and the metadata associated with the forgery. Because of this, any new forgery function capable of returning the forged image along with its manipulation map –a pixel-wise map locating the forgery inside the image– can be easily added to the library to generate \textit{Simple} or \textit{Compound} tampering figures.

\section{RECOD Scientific Image Integrity Dataset - RSIID}
\label{sec:forgery-data}
In addition to the library we introduced in the previous section, we created a dataset to serve as future benchmark for the area. For that, we selected the most frequent retracted types of images from the biomedical area. For this, we followed the orientation of Bucci \cite{bucci2018automatic} and Bik et al. \cite{Bik2016prevalence} that report a high image manipulation rate on images from Western Blot techniques and Microscopy imagery. With this in mind, we downloaded real scientific images collected from diverse sources to apply the forgeries.

To avoid any legal aspect of creating manipulated images and aiming to publish the dataset with a common creative license, we only downloaded data available under public domain\footnote{https://creativecommons.org/publicdomain/zero/1.0 (Last access May, 2021)} (PD) or common creative attributed\footnote{https://creativecommons.org/licenses/by/4.0 (Last access May, 2021)} (CC-BY) licenses. These license allow us to remix, transform, and reuse the images without asking for the author's authorization.

We use the following data source to gather the image collection:
\begin{enumerate}
    \item\textbf{Broad Bioimage Benchmark Collection\footnote{https://bbbc.broadinstitute.org (Last access May, 2021)} (BBBC), \label{BBBBC}}: \\
     A collection of freely downloadable microscopy image sets. From this source, we selected the datasets BBBC038, BBBC039, and BBBC019. The first two are dedicated to segmented nuclei images and have object-mask --that are needed for forgeries at object-level--. The last dataset (BBBC019) is dedicated to cell migration which we use for Overlap forgeries.
    \item \textbf{PubMed Central (PMC)\footnote{https://www.ncbi.nlm.nih.gov/pmc (Last access May, 2021)}}:\\
   PMC is a free article archive of biomedical and life sciences. To download each figure from this repository, we use an API\footnote{https://www.ncbi.nlm.nih.gov/pmc/tools/openftlist (Last access May, 2021} available by PubMed, in which we could select images that only have PD or CC-BY licenses. We choose to include published western blots images. To include the western blots to the source dataset, after downloading the PMC figures, we manually extracted the panels that had western blots for each figure. We based the templates images used for creating \textit{Compound Figures} based real figures retrieved from this repository.
    \item \textbf{TNBC \cite{Naylor2017}}:\\
    This dataset, announced in Naylor et al. \cite{Naylor2017}, was designed aiming at nuclei segmentation of cells by deep neural networks. Therefore, this dataset has a high-quality object map that we make use of to assists the forgery operations at object-level.
\end{enumerate}

Table \ref{tab:source-dataset} shows the number of source figures for each collection by type (Microscopy or Western blot) pointing if they have object-mask annotation, and the number of template images created based on the figures from PMC.

\begin{table}[]
\caption{RSSID Source Image and Template Collection}
\centering
\resizebox{0.4\textwidth}{!}{%
 \begin{threeparttable}
\begin{tabular}{|c|S[table-format = 5.0, table-number-alignment = center]|S[table-format = 5.0, table-number-alignment=center]|c|}
\hline
\multicolumn{4}{|c|}{\textbf{Source Image Dataset}} \\ \hline
\multicolumn{1}{|c|}{\textbf{Collection}} & \multicolumn{1}{c|}{\textbf{Microscopy}} & \multicolumn{1}{c|}{\textbf{Western Blot}} & \multicolumn{1}{c|}{\textbf{Object-Mask}} \\ \hline
BBBC039 & 800 & 0 & \ding{51} \\ \hline
BBBC038 & 552 & 0 & \ding{51}\\ \hline
BBBC019 & 165 & 0 & \ding{55} \\ \hline
TNBC & 50 & 0 & \ding{51}\\ \hline
PMC & 382 & 1009 & \ding{55} \\ \hline
\multicolumn{4}{|c|}{\textbf{Compound Figures Template}} \\ \hline
 \multicolumn{2}{|c|}{\begin{tabular}[c]{@{}c@{}}\textbf{Template Source\tnote{*}}\end{tabular}}   & \multicolumn{2}{c|}{\begin{tabular}[c]{@{}c@{}} \textbf{Templates}\end{tabular}} \\ \hline
 \multicolumn{2}{|c|}{\begin{tabular}[c]{@{}c@{}} PMC \end{tabular}}   & \multicolumn{2}{c|}{\begin{tabular}[c]{@{}c@{}} 321\end{tabular}} \\ \hline
\end{tabular}%
\begin{tablenotes}
   \item[*] Templates created based on the figures from the collection.
  \end{tablenotes}
 \end{threeparttable}
}
\label{tab:source-dataset}
\end{table}

\subsection{Dataset Construction}
\begin{figure}
        \centering
        \includegraphics[width=0.45\textwidth]{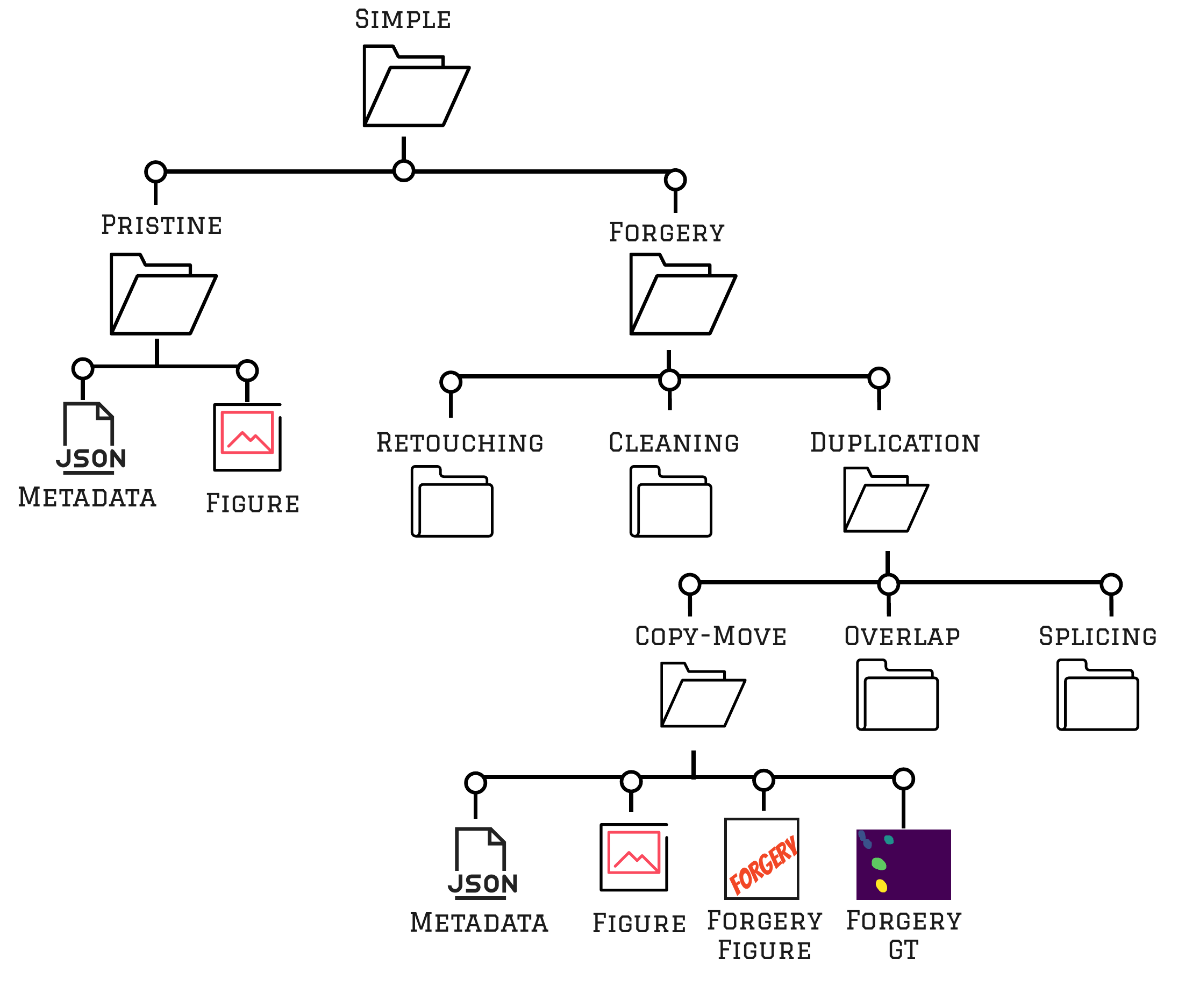}
        \caption{Organization of \textit{Simple} forgery images in the dataset.}
        \label{fig:simple-tampered-data}
\end{figure}

\begin{figure}
        \includegraphics[width=0.45\textwidth]{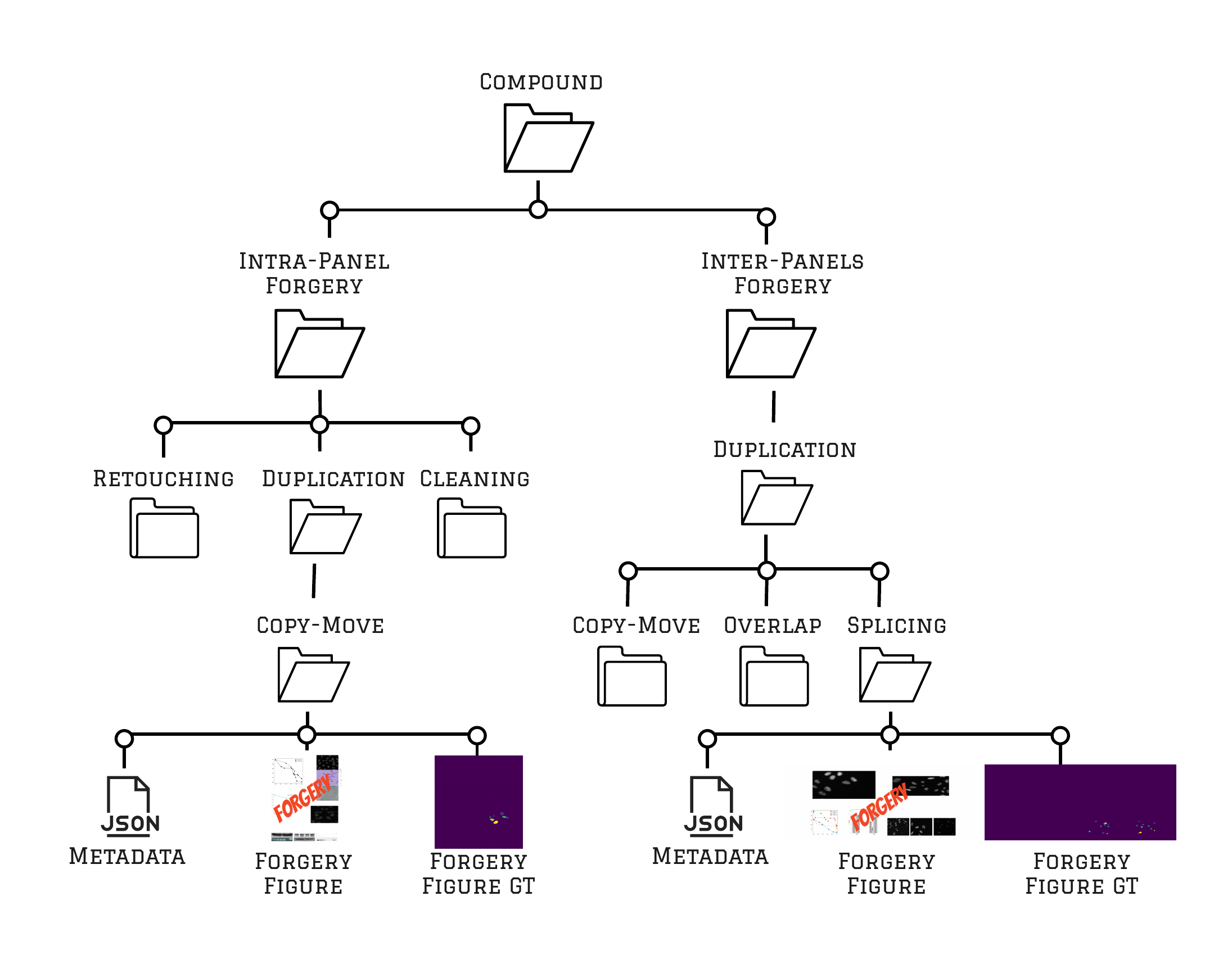}
        \caption{Organization of \textit{Compound} forgery images in the dataset}
        \label{fig:compoud-tampered-data}
\end{figure}

While designing the dataset, we project it to evaluate a forensic tool in different tasks with different complexities. Because of this, the dataset is organized so that a user can easily find the data and its annotation for each forgery modality.
Thus, we divided the dataset into two types of figure complexities: \textit{Simple} and \textit{Compound}.
\begin{enumerate}

\item \textit{{Simple Scientific Figures:}} Figures with this complexity are represented by a single experiment image, (e.g., Figure \ref{fig:retouching}a). To include a tampering figure in this complexity type, we forge an original figure using Retouching, Cleaning, or Duplication techniques, implemented in our library. To avoid unexpressive forgeries, we only included doctored figures in the dataset that have at least 500 manipulated pixels. In addition to the tampered figures, we also reserved a pristine directory in which we include the original images, so that a user can easily evaluate false positives. Figure \ref{fig:simple-tampered-data} illustrates the organization of \textit{Simple} figures in our dataset.

\item \textit{{Compound Scientific Figures:}}
This complexity type is described in Section \ref{subsec:realist-scifigs} and depicted in Figure \ref{fig:compound-creation-pipeline}.

We divided the \textit{Compound Figures} into two types of tampering: \textit{Intra-Panels} and \textit{Inter-Panels}.
\begin{enumerate}
    \item \textit{Intra-Panels} are forgeries that are present in just one panel of the figure. To create this tampering type, we add a \textit{Simple} forgery as one of the figure's panels.  Forgeries that need more than one source image (e.g., splicing) or that generate more than one doctored figure (e.g., overlap) were not included in this modality.
  \item \textit{Inter-Panel} are forgered figures that have two or more panels involved in the manipulation process. This modality aims to evaluate duplications among two or more panels within the same figure. These duplications can be at object-level, region-level, or panel-level. At object-level, the objects from a donor panel are copied into a host, using splicing operation. At region-level, an overlap forgery operation creates two panels with overlapping areas. At panel-level, the entire panel is duplicated with or without post-processing (e.g., retouching, cleaning, or geometric transformations).
\end{enumerate}

For each \textit{Compound} figure, we generated the three levels of indicative letters verbosity, as described in Section \ref{subsec:realist-scifigs}.
Figure \ref{fig:compoud-tampered-data} illustrates the organization of \textit{Compound} figures in our dataset.

To assist machine learning forensics techniques, we further divided the dataset into training/test sets. Tables \ref{tab:simple-dataset} and \ref{tab:compoun-dataset} express the number of manipulated figures included in each modality. Note that, overlap forgery appears only in the test set, since this modality is similar to the copy-move, and this protocol will force the generalizability of a forensic tool among the methods.


\end{enumerate}

\begin{table}[]
\caption{Number of Simple Figures per modality in the dataset}
\centering
\resizebox{0.5\textwidth}{!}{%
\begin{tabular}{|c|c|S[table-format = 5.0, table-number-alignment = center]|S[table-format = 5.0, table-number-alignment=center]|}
\hline
\multicolumn{4}{|c|}{\textbf{Simple}} \\ \hline
\multicolumn{2}{|c|}{\multirow{2}{*}{\textbf{Modality}}} & \multicolumn{1}{c}{\multirow{1}{*}{\textbf{Train}}} & \multicolumn{1}{|c|}{\multirow{1}{*}{\textbf{Test}}} \\ \cline{3-4} 
\multicolumn{2}{|c|}{} & \textbf{Number of Figures} & \textbf{Number of Figures} \\ \hline
\multicolumn{2}{|c|}{Source of Forgery Figures} & 1932 & 991 \\ \hline
\multicolumn{2}{|c|}{Pristine} & 1932 & 991 \\ \hline
\multirow{4}{*}{Duplication} & Copy-Move & 3761 & 1629 \\ \cline{2-4} 
 & Splicing & 604 & 274 \\ \cline{2-4} 
 & Overlap & 0 & 660 \\ \cline{2-4} 
 & \textbf{Total} & 4365 & 2563 \\ \hline
\multirow{3}{*}{Cleaning} & Inpainting & 275 & 117 \\ \cline{2-4} 
 & Brute-force & 961 & 412 \\ \cline{2-4} 
 & \textbf{Total} & 1232 & 529 \\ \hline
\multirow{3}{*}{Retouching} & Blurring & 961 & 414 \\ \cline{2-4} 
 & Contrast & 966 & 415 \\ \cline{2-4} 
 & \textbf{Total} & 1927 & 829 \\ \hline
\multicolumn{2}{|c|}{\textbf{Total of Figures}} & 9456 & 4912 \\ \hline
\end{tabular}%
}
\label{tab:simple-dataset}
\end{table}

\begin{table}[]
\caption{Number of Compound Figures per modality in the dataset}
\centering
\resizebox{0.5\textwidth}{!}{%
\begin{tabular}{|c|c|c|S[table-format = 5.0, table-number-alignment = center]|S[table-format = 5.0, table-number-alignment=center]|}
\hline
\multicolumn{5}{|c|}{\textbf{Compound}} \\ \hline
\multicolumn{3}{|c|}{\multirow{2}{*}{\textbf{Modality}}} & \multicolumn{1}{c}{\multirow{1}{*}{\textbf{Train}}} & \multicolumn{1}{|c|}{\multirow{1}{*}{\textbf{Test}}} \\ \cline{4-5} 
\multicolumn{3}{|c|}{} & \textbf{Number of Figures} & \textbf{Number of Figures} \\ \hline
\multicolumn{3}{|c|}{Source of Forgery Figures} & 1932 & 991 \\ \hline
\multirow{4}{*}{Inter-Panel} & \multirow{4}{*}{Duplication} & Copy-Move & 9516 & 4094 \\ \cline{3-5} 
 &  & Splicing & 604 & 274 \\ \cline{3-5} 
 &  & Overlap & 0 & 660 \\ \cline{3-5} 
 &  & \textbf{Total} & 10120 & 5028 \\ \hline
\multirow{8}{*}{Intra-Panel} & \multirow{2}{*}{Duplication} & Copy-Move & 3761 & 1629 \\ \cline{3-5} 
 &  & \textbf{Total} & 3761 & 1629 \\ \cline{2-5} 
 & \multirow{3}{*}{Cleaning} & Inpaiting & 275 & 117 \\ \cline{3-5} 
 &  & Brute-Force & 957 & 412 \\ \cline{3-5} 
 &  & \textbf{Total} & 1232 & 529 \\ \cline{2-5} 
 & \multirow{3}{*}{Retouching} & Blurring & 961 & 414 \\ \cline{3-5} 
 &  & Contrast & 966 & 415 \\ \cline{3-5} 
 &  & \textbf{Total} & 1927 & 829 \\ \hline
\multicolumn{3}{|c|}{\textbf{Total of Figures}} & 17040 & 8015 \\ \hline
\end{tabular}%
}
\label{tab:compoun-dataset}
\end{table}

\section{Copy-Move Forgery Detection Proposed Metric}
\label{sec:metric}

Popular metrics used on Copy-Move Forgery Detection (CMFD) (e.g., F1-score and Precision) make use of True Positive (TP), False Negative (FN), False Positive (FP), and True Negative (TN) detection concepts at pixel-level, as described in Table \ref{tab:confusion-matrix}. 

\begin{table}[]
\centering
\caption{Confusion Matrix Copy-Move Forgery Pixel Level}
\label{tab:confusion-matrix}
\resizebox{0.5\textwidth}{!}{%
\begin{tabular}{c|c|c|c|}
\cline{3-4}
\multicolumn{2}{c|}{\multirow{2}{*}{}} & \multicolumn{2}{c|}{\textbf{Predicted - Detection Map}} \\ \cline{3-4} 
\multicolumn{2}{c|}{} & Positive (Suspect Pixels) & Negative (Non-Suspect Pixels) \\ \hline
\multicolumn{1}{|c|}{\multirow{2}{*}{\textbf{Ground-Truth}}} & Positive ( Tampered Pixel) & TP & FN \\ \cline{2-4} 
\multicolumn{1}{|c|}{} & Negative (Pristine Pixel) & FP & TN \\ \hline
\end{tabular}
}
\end{table}

\begin{equation}
\mbox{F1-score} = \frac{2 TP}{2 TP + FN + FP}    
\end{equation}

\begin{equation}
\mbox{Precision} = \frac{TP}{TP + FP}    
\end{equation}

\begin{figure}
        \centering
        \includegraphics[width=0.4\textwidth]{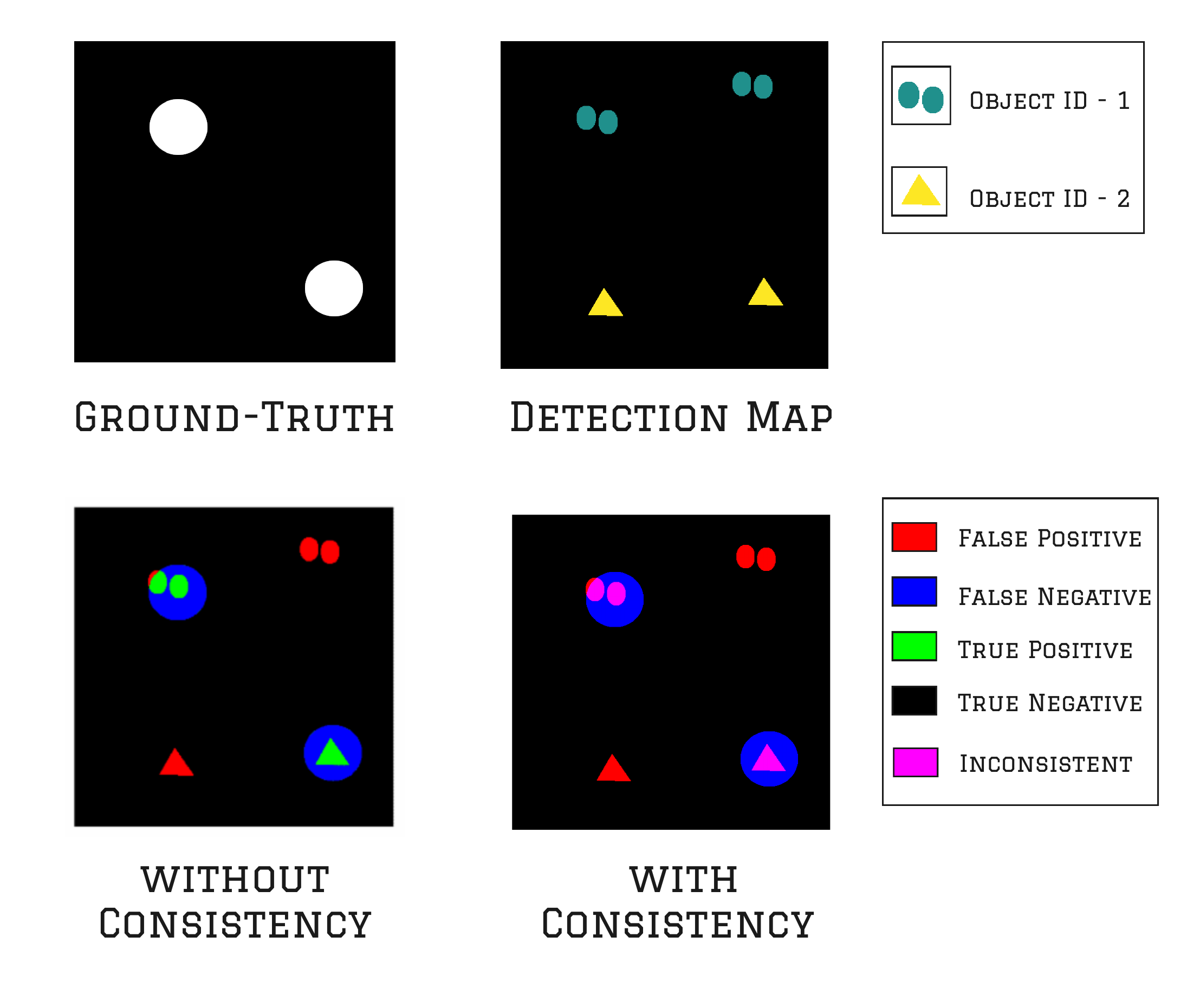}
        \caption{ Example of an inconsistent detection map. Although the detection map has two overlapping regions with the ground-truth, each object and its copy ---indicated by the detection map--- does not intersect simultaneously with the ground-truth.}
        \label{fig:metric-explanation}
\end{figure}

As a drawback, these metrics cannot assert if both regions of a copy-move (the source and its copy) are in the ground-truth, since there is no consistency check. Because of this, some contradiction might occur during the evaluation. 
For instance, Figure \ref{fig:metric-explanation}  illustrates a detection map that has an inconsistent detection in which the copied objects with $id=1$ and $id=2$ are inconsistent with the ground-truth. For both objects, only the source or its copy overlaps with the ground-truth, which is inconsistent, since both (object and its copy) are expected to be included during detection. However, when evaluating this detection map with traditional true positive score, both regions would be considered as true positive hit.

To mitigate this drawback, this section introduces a new metric that takes advantage of the enriched pixel-wise ground-truth of the dataset. The proposed metric is a variation on how to consider a pixel as true positive in a detection map named as the Consistent True Positive score (\textit{CTP}) and defined as: \newline
Given a ground-truth map $GM$ with $n\geq1$ manipulated regions, a detection map $DM$ with $m\geq0$ copy-pasted regions, each one of the $n$ regions included in $GM$ has $c_{gm}\geq2$ connected components (the source object and all its copies), and each region in $DM$ has $c_{dm}\geq1$ connected components. Let $R_{dm}$ be a detected region from $DM$ and $R_{gm}$ a tampered region indicated by the ground-truth. Also, let $p$ be a pixel from $DM$, such that $p \in R_{dm}$.
Thus, $p$ is a consistent true positive if exists $R_{gm} \in GM$, such that, at least two connected components from $R_{gm}$ intersects $R_{dm}$.

In other words, to consider a region $R_{dm}$ from the detection map as a consistent true positive, at least two components from  $R_{dm}$ (the source and at least one of its copies) have to intersect the ground-truth.

\begin{figure}
        \centering
        \includegraphics[width=0.4\textwidth]{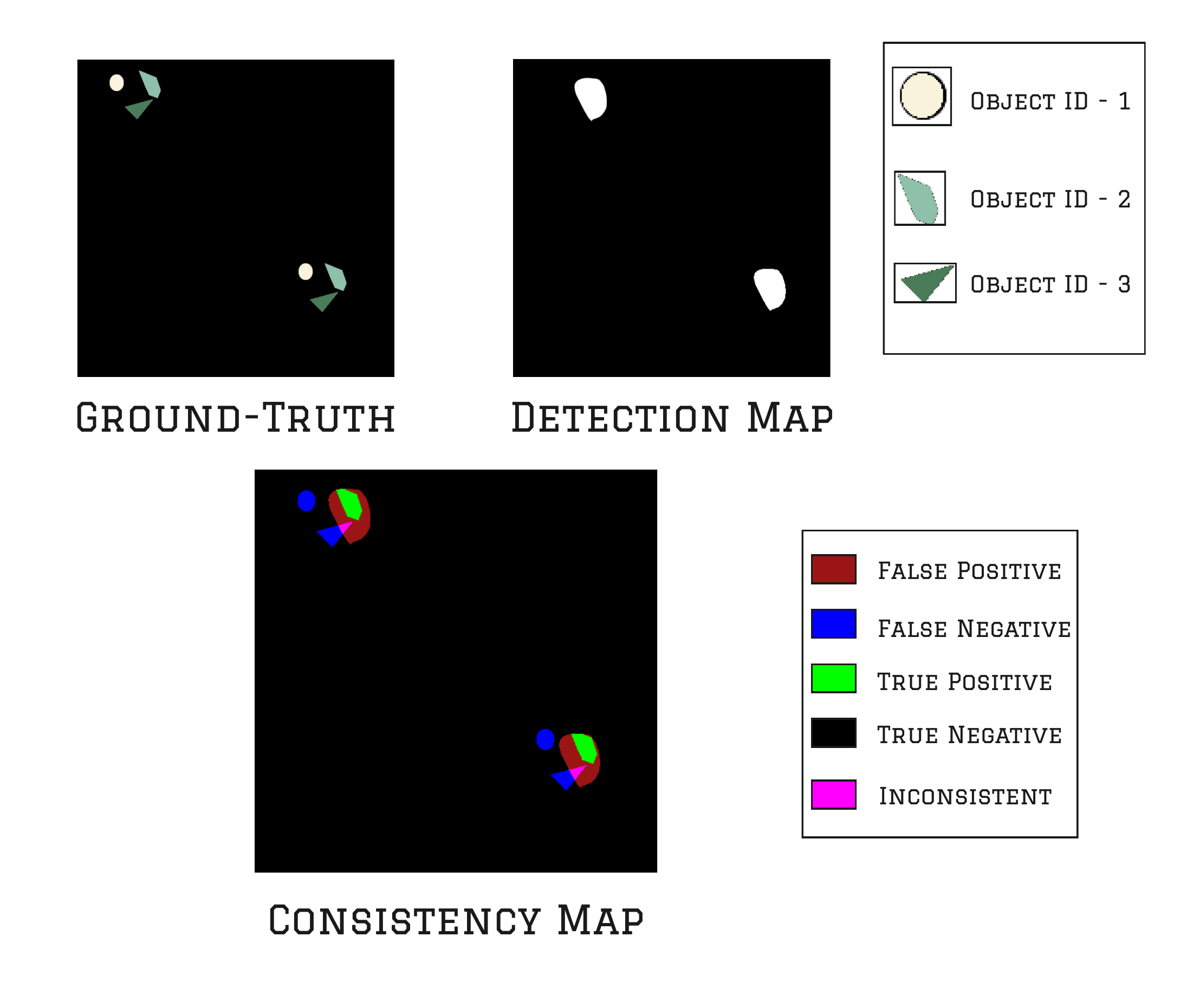}
        \caption{Example of \textit{CTP} for a detected region that overlap two or more objects from the ground-truth. The ground-truth present three copied objects. The detection map present two connected components with the same ID that overlaps more than one object from the ground-truth. Only the object with the larger area is considered as true positive, the others would be considered inconsistent.}
        \label{fig:metric-decision}
\end{figure}

As Figure \ref{fig:metric-decision} depicts, a region from the detection map can overlap with two or more region from the ground-truth. Given that the goal of $CTP$ is consistency, we only consider as $CTP$ the region of the ground-truth that has the maximum intersection area with the detected region. Hence, $CTP \leq TP$. As a result, $FN$ will have higher penalty on Precision and F1-score metrics, if calculated with $CTP$.

Thus, the equation of F1-score and Precision using \textit{CTP} become:
\begin{equation}
\mbox{F1-score}_{CTP} = \frac{2 CTP}{2 CTP + FN + FP}    
\end{equation}

\begin{equation}
\mbox{Precision}_{CTP} = \frac{CTP}{CTP + FP}    
\end{equation}



\section{Evaluating CMFD Methods}
\label{sec:eval}
Duplication of scientific images is one of the threats highly reported and studied in the literature \cite{Bik2016prevalence, rossner2004what, Wjst2021} which includes copy-move, a well-studied forgery in the digital forensic field. Although this field presents multiples CMFD solutions for natural images, we could not find any study that evaluates their performance in the scientific image domain. In this sense, to assist any future forensic method with a baseline, we investigated the performance of popular CMFD solutions on natural images applied to the \textsc{RSIID}. In addition, we checked the difference of F1-score using the proposed consistent true positive metric and the regular true positive one. For this, we choose the following CMFD methods:
\begin{enumerate}
    \item Efficient Dense-Field from Cozzolino et al. \cite{cozzolino2015efficient}. During the evaluation, we use the implementation of Ehret \cite{Ehret2018}. Ehret released two versions of \cite{cozzolino2015efficient} using Zernike and SIFT features. To distinguish this method from the others, we named them Zernike-PM and SIFT-PM, since these detectors use the PatchMatch algorithm \cite{barnes2009patchmatch} to match similar blocks contents.
    \item CMFD library implemented by Christlein et al. \cite{Christlein2012}. We selected SIFT and SURF methods from this library since the others were not efficient enough to be explored on such a large dataset. To distinguish them from the previous CMFD detectors, we named them SIFT-NN and SURF-NN since they use a regular approximate nearest-neighbor approach to match similar blocks. 
    \item Busternet from Wu et al. \cite{wu2018eccv}. This method is a deep neural architecture for CMFD. 
    During the evaluation, we use the pre-trained version of the model released by Wu et al. \cite{wu2018eccv}. 
\end{enumerate}

To evaluate SIFT-PM, Zernike-PM, SIFT-NN, and SURF-NN using $CTP$, we modified their implementation, including a routine that assigns each detected object and its copies a unique ID. For the sake of reproducibility, we released the methods source-code with this modification in the same repository of RSIIL. On the other hand, to evaluate Busternet, we normalized its output [0,255], then binarized all pixels greater than 100 to 1, otherwise 0. As Busternet is based on neural networks, we could not find an explainable methodology that would track the matching among different objects and their copies. Thus, to the $CTP$ metric, all detected and ground-truth objects are set with the same $ID=1$. Consequently, $CTP$ would not be able to properly check inconsistencies on figures with more than one tampered object for Busternet's output; however, $CTP$ is still valid and useful to check if Busternet's output overlaps with two or more connected components from the ground-truth.

\begin{figure*}
        \centering
        \subfloat[CMFD Simple Figure Evaluation. The best method is Busternet because its polygon contains all the others, indicating that it performs better for all operations. The shrinking polygons area from $\mbox{F1-score}_{TP}$ to $\mbox{F1-score}_{CTP}$ indicates that all methods show inconsistencies in their detection map.\label{fig:simple-eval}]{%
        \includegraphics[width=0.9\textwidth]{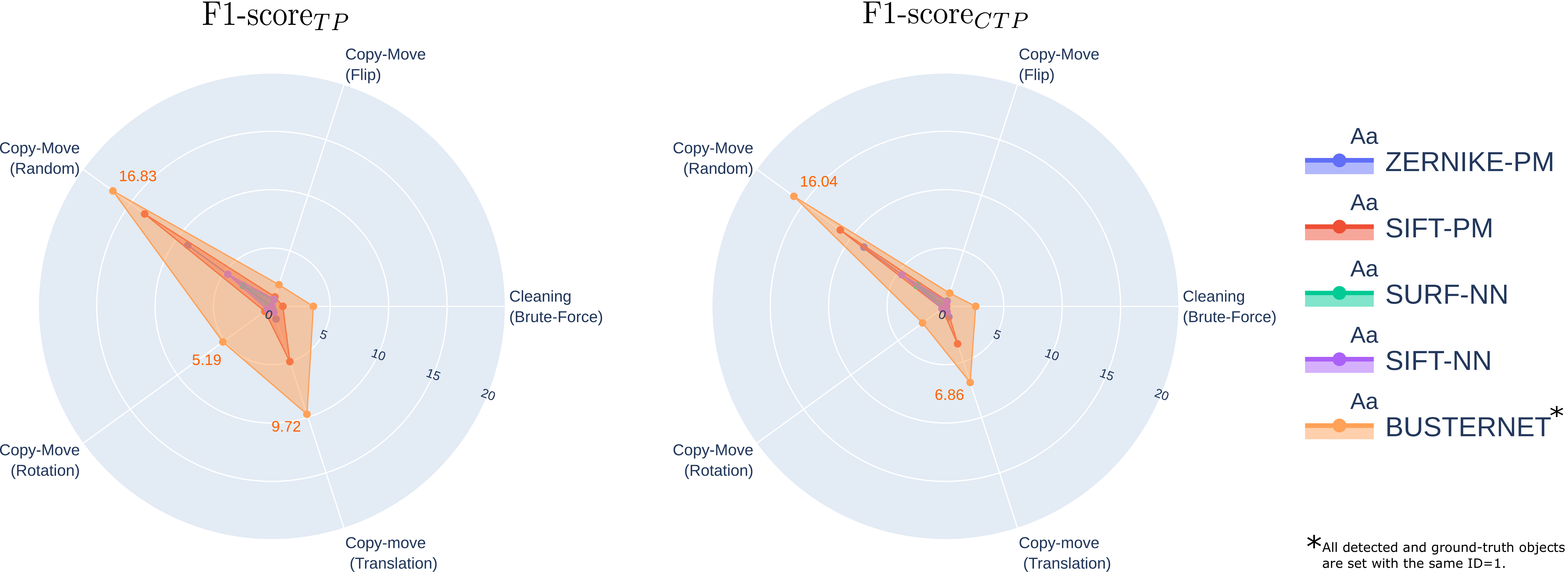}
        }\hfill
        \subfloat[CMFD Inter-Panel Figure Evaluation on different levels of indicative letters verbosity. In this plot, each method fares differently for each modality. The polygons from SIFT-NN and SURF-NN have larger area than the others methods, indicating that they are robust to more operations than the others methods. The shrinking polygons area from $\mbox{F1-score}_{CTP}(V1)$ to $\mbox{F1-score}_{CTP}(V3)$ indicates the higher the caption Level, the lower is the method’s effectiveness. \label{fig:inter-eval}]{%
        \includegraphics[width=\textwidth]{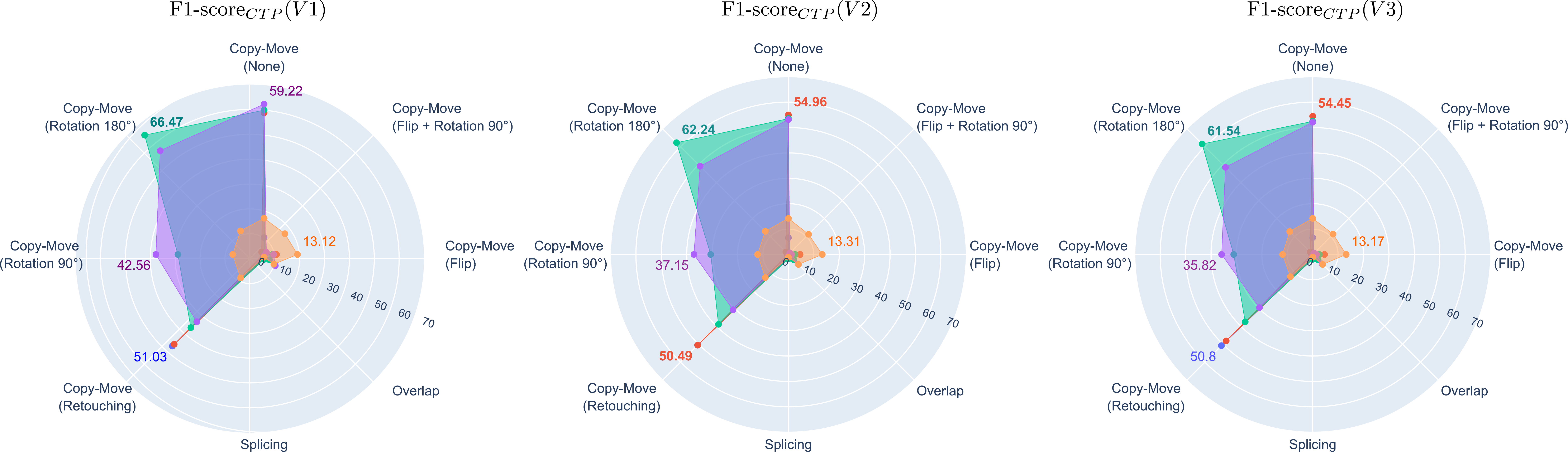}}
        \hfill
         \subfloat[CMFD Inter-Panel Figure Evaluation on different levels of indicative letters verbosity. All methods show low performance and concentrate in the center of the radar, indicating that this is a challenging modality.\label{fig:intra-eval}]{%
        \includegraphics[width=\textwidth]{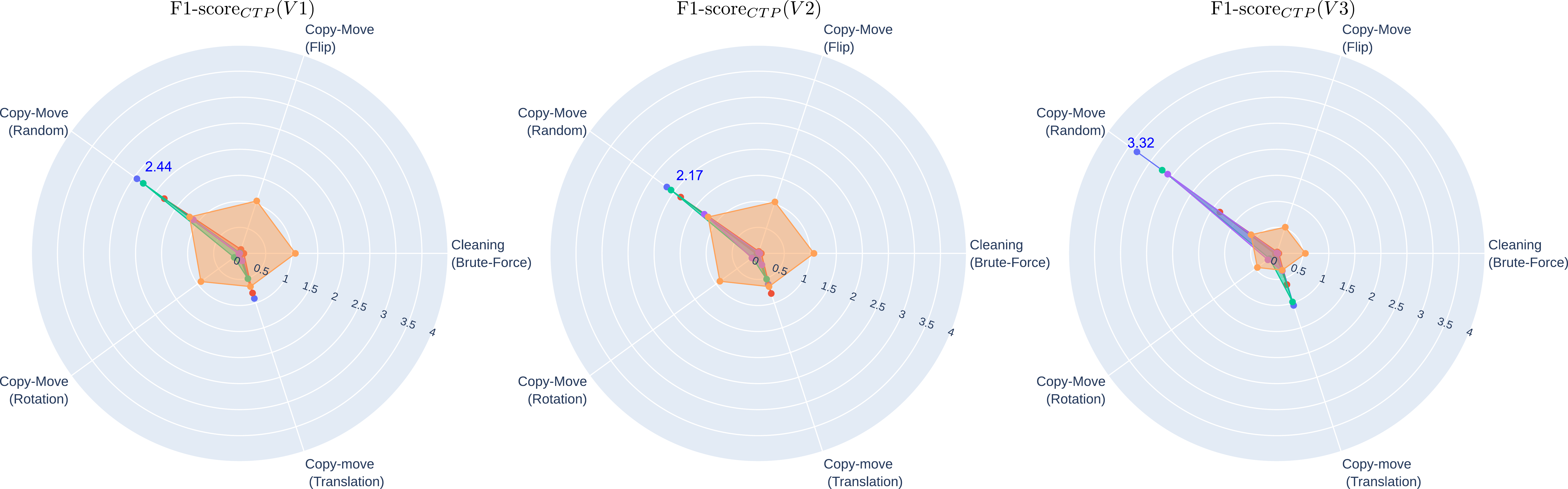}
        }
         \caption{Evaluation Baseline Results. Inside the parenthesis of each copy-move modality, there is the transformation used during the copy-move forgery. All F1-scores presented in this figure are normalized $[0,100]$. The best result for each duplication modality is indicated with the color of the respective detector. (a) Result for Single Figure Evaluation using $\mbox{F1-score}_{TP}$ and $\mbox{F1-score}_{CTP}$. (b) Result for Inter-Panel Figure Evaluation using $\mbox{F1-score}_{CTP}$ across all levels of indicative letters verbosity, indicated by the number in its subtitle (i.e., V1 for verbosity Level 1). (c) Result for Intra-Panel Figure Evaluation using $\mbox{F1-score}_{CTP}$ across all levels of indicative letters verbosity.}
        \label{fig:evaluation}
\end{figure*}

\begin{figure*}
\centering
 \large{Duplication detection output per modality}
 
 \vspace{0.3cm}
    \includegraphics[width=\textwidth]{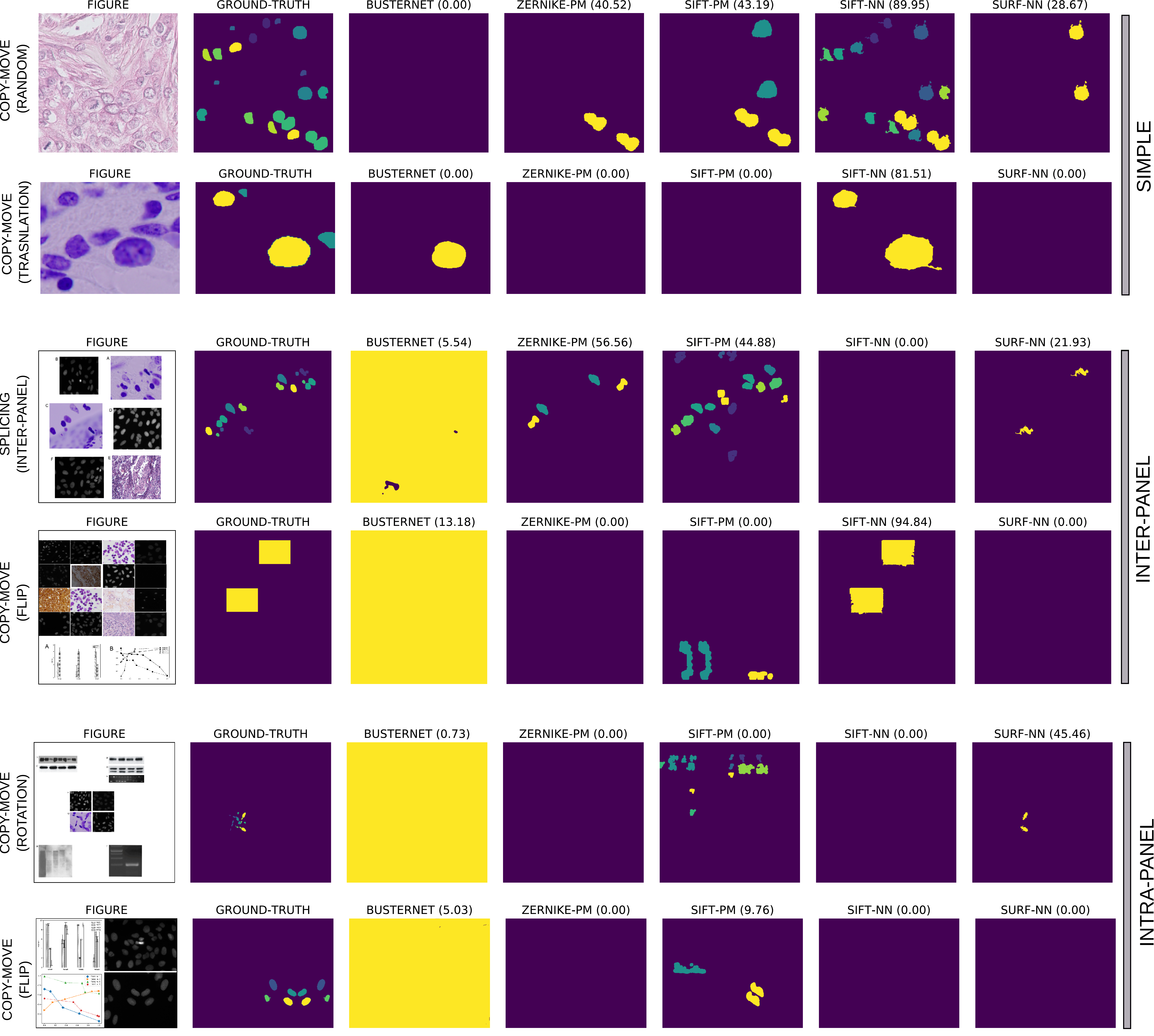}
         \caption{Comparative duplication detection output per modality. The purple color represents a pristine/non-suspect region, and each other color in the ground-truth and detection maps represents a different ID assigned to each object and its copies. Inside the parenthesis of each method, we insert the $\mbox{F1-score}_{CTP}$ metric normalized into [0, 100]. All compound figures are with level 1 of indicative letters verbosity.}
    \label{fig:vis-results}
\end{figure*}

As a baseline approach, the evaluation protocol consists of running all methods without any training or fine-tuning and measuring their output with $\mbox{F1-score}_{CTP}$. During the evaluation, we use all figures from the test set applicable for CMFD (i.e., images with duplicated areas within the same image).
We group the baseline results into \textit{Simple} and \textit{Compound} scientific figures, which were divided by modalities. All copy-move modalities presented in Figures \ref{fig:evaluation} and \ref{fig:vis-results} can also include scaling. Since scaling cannot be applied alone, we did not indicate when this operation is combined with others.


Figure \ref{fig:evaluation} presents a radar graph visualization in which the forgery modalities are arranged in the radius axes. Each CMFD methods' result is represented with a different color in the radar char. In this visualization, we insert the score of each method along the modality axis (e.g., copy-move with flip) which start from the radar center (score zero) until its border (highest score); thus, as farther a method point (color point) is from the center as better is the method for the axis copy-move modality. After inserting all points of a method for each copy-move modality, we connected those points resulting in a polygon. The larger the polygon area, the better the method's robustness among different forgery modalities. Also, comparing each detector's robustness to the operations, this type of visualization helps identifying possible complementary behaviors among different methods.
As an example, consider Figure \ref{fig:simple-eval} left panel. In this case, we have five modalities being compared (e.g., Copy-Move with Flip, Cleaning with Brute-Force, and Copy-Move with Translation). This char shows the results of five methods represented by each different polygon color (see legend on the right of the figure). The best method in this figure is Busternet (in orange) while the two worse methods (SURF-NN and Zernike-PM) are in superposition at the center (smaller areas). 
In the following, we discuss the forgery evaluation for each modality.

\subsection{Simple Figure Forgery Baseline}
In this modality, we tested the chosen methods on \textit{Simple} figures, forged with Cleaning (Brute-Force) and Copy-Move.
Although the chosen CMFD detectors have high efficacy on natural image benchmarks, their performance drastically decrease when applied to our scientific dataset. As Figure \ref{fig:simple-eval} shows, the best CMFD method in the Simple Figure Evaluation was Busternet \cite{wu2018eccv}, despite its modest scores.

For this modality, we also compare each methods' performance between F1-score using $TP$ and $CTP$. We notice a difference in these scores for all methods, represented by the area reduction from their polygon chars, indicating the existence of copy-move inconsistencies on their detection maps, which is also depicted in Figure \ref{fig:vis-results}.
The second row of this figure shows an example of an inconsistent detection map, in which Busternet activates just one of the connected components involved in the manipulation, resulting in $\mbox{F1-score}_{CTP}=0$. On the other hand, in the same row, SIFT-NN detects both regions (object and copy), resulting in $\mbox{F1-score}_{CTP}=81.51$.

\subsection{Inter-Panel Figure Baseline}
We evaluate the \textit{Inter-Panel} tampered figures for all indicative verbosity levels in Copy-Move (at panel-level), Splicing, and Overlap forgeries.
Figure \ref{fig:inter-eval} shows the result for the \textit{Inter-Panel} forgery evaluation using $\mbox{F1-score}_{CTP}$. In this modality, the radar visualization allowed us to notice some complementary performance among the chosen detectors. For instance, SURF-NN and Zernike-PM show a complementary behavior to copy-move with rotation and retouching. We believe that this complementary aspect indicates that a fusion/ensemble technique might enhance their individual robustness.
For the \textit{Inter-Panel} scenario, the flipped copy-move, Splicing, and Overlap forgery showed to be the most challenging forgeries. In addition, the indicative letters are shown to have a perceptible impact in this scenario, reducing by up to seven points from level 1 to level 3 for some detectors.
Althouth Busternet achieves the best performance in \textit{Simple} Figure modality, when applied to compound figures, it leads to a higher false-positive rate, as depicted in Figure \ref{fig:vis-results} by activating the entire image.

We also noticed that graphs and indicative letters are the most common causes of false alarms in the \textit{Compound} Figures scenario, as illustrated by the third, fourth, and fifth rows of Figure \ref{fig:vis-results}, which SIFT-PM wrongly activates letters and graphs regions.

These findings help us to see where researchers should focus on when dealing with the scientific image forgery detection problem.

\subsection{Intra-Panel Figure Baseline}
We evaluate the \textit{Intra-Panel} tampered figures for all levels of indicative verbosity in Cleaning (Brute-Force) and Copy-Move (at object-level).

As presented by Figure \ref{fig:evaluation}, this is the most challenging scenario, in which the detectors scored lower than four on $\mbox{F1-score}_{CTP}$ for all evaluated operations. A possible explanation for this is the lower percentage number of doctored pixels in these figures than in other modalities. The detectors' low performance does not allow us to measure the impact of verbosity levels in the figures properly. However, as Figure \ref{fig:vis-results} shows, graphs and indicative letters would also be one cause of false alarms in this modality.

\section{Conclusions}
\label{sec:conclusions}
 In addition to the daunting scenario of fraud in science ---due to the increase of image misconduct cases---, there is a legal issue related to copyrights and judicial aspects that prevents one from creating a large collection of fraudulent scientific images, even for an in-depth forensic study to benchmark and drive the development of appropriate detection methods.

Therefore, this work introduced a library and a dataset to assist the scientific integrity and forensic community to overcome this legal hurdle. We believe that by presenting a large dataset to the forensic community, we are fostering the development of more complex and robust detection tools (e.g., AI-based models).

The proposed library implements the most common image manipulation forgeries described by scientific integrity researchers. Also, it is extendable to more complex tampering operations. As a special feature, the library generates an enriched ground-truth addressing all regions affected before and after applying a tampering function, assigning a unique ID for the regions involved (when applicable). 
Using this library on creative common scientific images, we created a dataset with 39,423 manipulated figures freely available.

Leveraging the dataset's enriched ground-truth, we proposed a metric that avoids inconsistent detection during CMFD evaluation. Using this metric, we evaluate popular CMFD methods on our dataset. Although we choose high-cited and effective CMFD tools for natural images, all solutions presented a lower performance when transferred to the scientific image domain. This is not a fault of such algorithm as they were not designed for this specific setup. However, these findings show an important lack of methods and a tremendous research opportunity for new specialized detectors aiming at finding forgeries in scientific-related images. In addition, we notice that some of the chosen algorithms present complementary performance and might benefit from a fusion approach.

Notwithstanding the large size and diversity of the proposed dataset, we believe that science will report more sophisticated tampering operations in the near future, as warned by \cite{qi2020emerging}. Thus, we are also concerned about more issue-less and freely available scientific integrity datasets with complex, enhanced, and realistic tampering modalities, aiding the design of more robust detectors. 

Therefore, as future work, in addition to investigating robust forensic solutions using AI-based or fusion-based methods, we believe that studies on automated realistic scientific forgeries would also assist the forensic community in fighting scientific misconduct. 
Furthermore, we believe that the detailed pixel-wise ground-truth of RSIID opens a research opportunity to explore eXplainable AI solutions that might assist analysts on sensitive cases, such as misconduct investigations.


\section*{Acknowledgments}
This research was supported by São Paulo Research Foundation (FAPESP), under the thematic project \textit{DéjàVu}, grants 2017/12646-3 and 2020/02211-2.



\bibliographystyle{IEEEtran}
\bibliography{IEEEabrv,\theabsdir/../references}

\end{document}